# An Adaptive and Scalable ANN-based Model-Order-Reduction Method for Large-Scale TO Designs

Ren Kai Tan, Chao Qian, Dan Xu, Wenjing Ye*

Department of Mechanical and Aerospace Engineering, The Hong Kong University of Science and Technology, Clear Water Bay, Kowloon, Hong Kong

Department of Computer Science and Engineering, The Hong Kong University of Science and Technology, Clear Water Bay, Kowloon, Hong Kong

*Corresponding author: Wenjing Ye

Email: mewye@ust.hk

**Acknowledgement and Funding**

This work is supported by the Hong Kong Research Grants under Competitive Earmarked Research Grant No. 16206320.

**Abstract**

Topology Optimization (TO) provides a systematic approach for obtaining structure design with optimum performance of interest. However, the process requires numerical evaluation of objective function and constraints at each iteration, which is computational expensive especially for large-scale design. Deep learning-based models have been developed to accelerate the process either by acting as surrogate models replacing the simulation process, or completely replacing the optimization process. However, most of them require a large set of labelled training data, which are generated mostly through simulations. The data generation time scales rapidly with the design domain size, decreasing the efficiency of the method itself. Another major issue is the weak generalizability of most deep learning models. Most models are trained to work with the design problem similar to that used for data generation and require retraining if the design problem changes. In this work a scalable deep learning-based model-order-reduction method is proposed to accelerate large-scale TO process, by utilizing MapNet, a neural network which maps the field of interest from coarse-scale to fine-scale. The proposed method allows for each simulation of the TO process to be performed at a coarser mesh, thereby greatly reducing the total computational time. Moreover, by using domain fragmentation, the transferability of the MapNet is largely improved. Specifically, it has been demonstrated that the MapNet trained using data from one cantilever beam design with a specific loading condition can be directly applied to other structure design problems with different domain shapes, sizes, boundary and loading conditions.

**Keywords**: deep learning, neural network, topology optimization, surrogate method.

## 1. Introduction

Structural design has always been playing an important role in engineering field from the design of industrial products such as aircraft components to the design of advanced materials like metamaterials. One of the common design methods especially in the field of mechanical engineering is topology optimization (TO) in which the material distribution is optimized systematically within a prescribed design domain to achieve a design objective while subjecting to certain design constraints. However, due to the repetitive evaluations of the objective function and constraints required during the TO process, which are typically carried out by numerical simulations such as Finite Element Method (FEM)-based analysis, the computational cost of the TO method could be prohibitively large for large-scale designs. For example, the design of an airplane foil with giga-hertz resolution required 8000 CPUs running simultaneously for days [1].

One of the effective solutions to accelerate the TO process and reduce the computational cost of large-scale designs is to speed up the large-scale simulation. Various methods have been developed over the years, starting with conventional reduced order methods used in early years [2]-[8] to rapidly developed deep learning methods such as artificial neural networks (ANN) in recent years. One of the advantages of ANNs is that once constructed, predictions from ANNs can be rapidly computed with the time scale on the order of milliseconds. This advantageous property of ANN has been utilized in large-scale analysis with the ANN serving as the surrogate model to replace time-consuming numerical simulations such as FEM calculations. ANN-based surrogate models have been widely adopted in the field of structural mechanics for the prediction of mechanical responses of structures. This implementation is seen in the works [9]-[13], which utilized either the regular NN or Convolutional Neural Network (CNN) for the prediction of mechanical fields such as stress/strain field of structures subjected to various loadings.

Another approach to speed up the design process is to apply deep learning models to directly predict the optimized or near-optimal structures, partly or completely skipping over the optimization process, which is seen in works [14]-[17]. In these works, deep learning models are used to predict from "structures" to "structures", essentially treating structure designs as images. For example, in the work by Sosnovik et. al. [14], a neural network is used to predict the optimized structure directly using the density field of the structure at intermediate steps of TO. These models are used as black boxes without requiring any prior knowledge associated with the design problem, and the computational time is reduced by skipping the design process. However, in order to train the model, many TO design solutions must be pre-produced, which in turn requires a large number of FEM calculations.

Despite the great interest and effort on the development of machine learning-based efficient TO methods, most existing methods/models suffer from two major shortcomings. The first one is the need of a large set of training data. With the required data mostly generated through time-consuming simulations such as FEM calculations, the time required for the generation of training data increases with the problem size, which is impractical for large-scale structural designs. Several methods have been proposed to reduce the number of training data. For example, in the work by Qian and Ye [18], only the designs in the early stage of optimization are used as training data. The data from the later stage of optimization are omitted since most of these designs are similar to those in previous iterations and therefore do not contribute much to the learning process of the machine learning model. With this approach, the number of training data can be reduced to a certain extent, but significant amount of expensive FEM calculations is still required to be performed. This might become a problem for large-scale TO designs, for example, those with giga-scale resolution [1], [19], as the computational resource required for the calculations is too high and may not be easily accessible. In another work by Chi et al. [20], an online updating scheme is proposed to avoid the long offline training, thereby avoiding the need for a large pool of pre-generated training data. Since an online training scheme is utilized, the model is required to be trained in real time for every new problem.

Another drawback in most of the existing works is the poor transferability of the trained models, with their prediction accuracy dropping significantly when applied to "unseen" settings. As a result, most existing models have rather narrow application scopes, difficult to generalize to problems with different settings particularly with different domain shapes and sizes. A main reason for the narrow application scope is that in those methods, the machine learning model maps the entire domain, which is represented as an image, and possibly boundary/ loading conditions to its corresponding output such as the stress field or the design solution. To accommodate for different domains and conditions, a large set of representative training data is required, which is challenging if not impossible to generate for large-scale designs.

One approach to reduce the number of training data and/or improve the transferability is to incorporate the physics into the machine learning process or machine learning models. An example of the former case is the physics-informed neural network (PINN) approach [21]-[24], which has attracted a great attention recently. In PINN, the physics is embedded in the loss function. As a result, very little or even no training data is needed. The trade-off is that the training time greatly increases because the complexity of the training, which is effectively a multi-objective optimization problem, increases. The latter case can be seen in the work of Wang et al. [25] in which the neural network maps the mechanical field of the initial design such as the displacement and/or the stress/strain field to the corresponding design solution. The input of the ANN model contains certain physics of the problem. Hence the model has a relatively strong generalization ability. It can predict design solutions of the same problem with different boundary conditions even though the model was trained only on one boundary condition.

However, the quality of the produced design solutions needs to be improved and the model seems to be only applicable for a fixed design domain.

In this work, we propose and develop an ANN-based model-order-reduction method to greatly reduce the computational cost of the expensive numerical evaluation of the objective and constraints and thus to speed up the TO process for large-scale designs. In particular, this method is scalable and adaptable to different problem settings including the changing domain size and shape without the need to retrain the deep learning model. In the method, the objective and/or constraints are evaluated on a coarser mesh instead of the original mesh using conventional methods such as the FEM. A neural network model, entitled as MapNet, is then used to map the coarse-scale field to the full-scale field. The major benefit of this approach is that the coarse-scale field, which is much less expensive to obtain, contains the physical information, such as the boundary and loading conditions. If boundary/loading conditions change, the coarse-scale field changes accordingly. Compared with ANN models that map the structure to its full-scale field, the training of MapNet, is easier and requires less full-scale data because the network only needs to learn the relationship between a coarse field and its corresponding fine field. This approach is not new and has been applied in several works previously [20], [26]. To improve the transferability and scalability, the idea of domain decomposition is utilized in this method. The problem domain is decomposed into a set of small subdomains or fragments, and the network is constructed to perform the mapping on each small subdomain instead of the entire problem domain. The predicted field of each subdomain is then combined to form the field of the original domain. A major advantage of this approach is that many different domains with varying sizes and shapes can potentially be decomposed into similar sets of subdomains/fragments, the MapNet trained with data from a specific design problem can be more easily transferred to different problems. Besides, the number of training data is increased because one data of the entire domain can be decomposed into many subdomain data and thereby increasing the accuracy of the network.

The paper is presented by first describing the design problems used for the demonstration of the performance of the proposed method. The detailed implementation of the proposed method is discussed next. In the section of results and discussions, the accuracy and efficiency of the method are demonstrated on various design problems having different design domains and boundary conditions, benchmarked with results obtained from conventional TO methods. Finally, the paper is concluded with a discussion of the possible future work.

## 2. Problem Statement

The performance of the proposed method is demonstrated on two benchmark design problems of TO methods, specifically the structural and thermal compliance minimization design problems. The structural compliance minimization design problem can be expressed mathematically in a discretized form as follows:

$$\min_{x}: C(x) = \frac{1}{2}\mathbf{U}^{\mathbf{T}}\mathbf{K}\mathbf{U}$$

$$\text{subjected to:} \quad \frac{V(x)}{V_0} = f$$

$$\mathbf{K}\mathbf{U} = \mathbf{F}$$

(1)

where $C(x)$ is the compliance of the structure, **U** is the displacement vector, **K** is the global stiffness matrix, **F** is the loading vector. $V(x)$ represents the volume of the structure and $V_0$ is the volume of the entire design domain. $f$ is the desired volume fraction. In this work, the volume fraction constraint is set as 0.4 for all cases, unless specifically mentioned. The design variable is the elementwise "density" $x$. In this work, two common TO methods, the Bi-directional Evolutionary Structural Optimization (BESO) and the Solid Isotropic Material with Penalization method (SIMP), are used to perform TO designs. The elementwise density is either 0 or 1 in BESO and a value between 0 and 1 in SIMP with 0 representing the void and 1 representing the solid phase of the structure respectively. The design objective is to optimize the topology, that is, the density, of structures within the design domain, so that the compliance of the structure is the minimal, while subjecting to a given loading and volume fraction. A variety of design cases with different domains, boundary and loading conditions are selected to demonstrate the transferability of the proposed method, starting with the classical 2D cantilever beam design as illustrated in Fig 1(a). The design domain is a square with the left boundary being fixed and a vertical distributed load applied at the upper right boundary of the domain. The second design problem is the same as the cantilever design except that instead of one distributed load, three distributed forces are applied at the centers of the top, the right and the bottom boundaries respectively as illustrated in Fig 1(b). Two more design cases, that is, the L-shaped beam and simple bridge designs shown in Fig 1(c) and Fig 1(d), are also considered. The two cases have distinct domain shapes and sizes. In addition, boundary and loading conditions are entirely different from those of cantilever design problems. For the beam design of (a), (b), and (c), the distributed load is applied on a square region of L/16 by L/16, while for the bridge design problem the load is distributed over the whole top boundary.

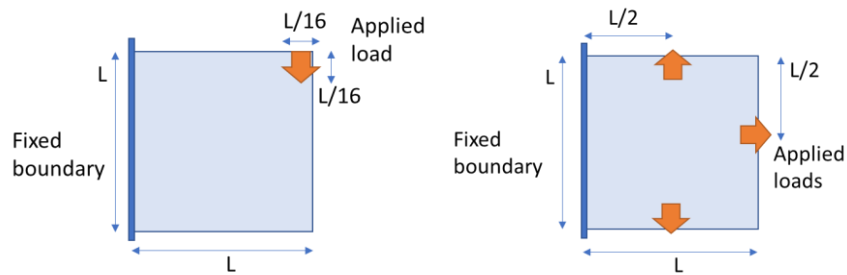

(a) Cantilever beam with load applied at upper right    (b) Cantilever beam with multiple loads

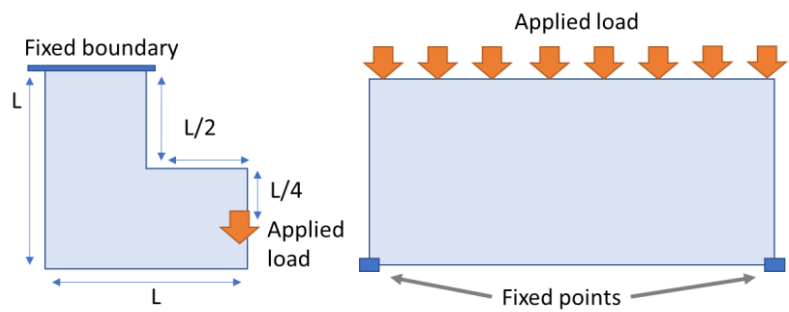

(c) L-shaped beam    (d) Bridge design

**Fig 1 Illustration of design domain for structural compliance minimization problem for (a) cantilever beam with load applied at the upper right corner, (b) cantilever beam with multiple loads, (c) L-shaped beam, (d) Bridge design**

In order to demonstrate the versatility of our method, the thermal compliance minimization problem is also solved using the proposed method. In this problem, the design objective is to minimize thermal compliance subject to a given thermal loading and boundary conditions. The mathematical model of the design problem is given as follows:

$$\min_{x}: C(x) = \mathbf{T}^T \mathbf{K}_c \mathbf{T}$$

$$\text{subjected to:} \quad \frac{V(x)}{V_0} = f$$

$$\mathbf{K}_c \mathbf{T} = \mathbf{F}$$

$$0 \leq x \leq 1$$

(2)

where C(x) is the objective function, $\mathbf{T}$ is the temperature field, $\mathbf{K}_c$ the conductivity matrix and $\mathbf{F}$ is the thermal loading including contributions from boundary heat flux and internal heat generation/loss. SIMP method is used for this problem and the design domain is a square. This design problem is based on that demonstrated in the work of Bendsoe et. al. [27], in which the design domain is isolated at all edges except for the heat sink with zero temperature located across the centre of the top boundary. The plate is subjected to distributed heating all over the plate. Two different boundary conditions are considered, with the first one having a small heat sink while the second having a large heat sink. The lengths of the heat sink in the two cases are illustrated in Fig 2.

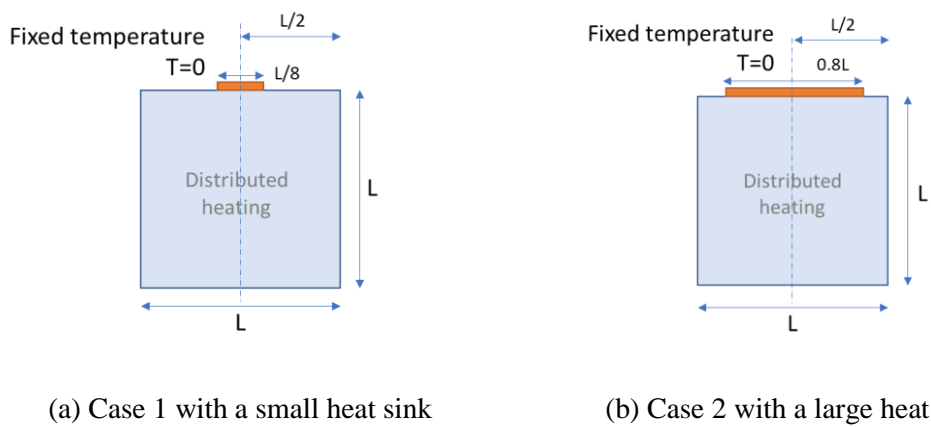

(a) Case 1 with a small heat sink     (b) Case 2 with a large heat sink

**Fig 2 Illustration of design domains of thermal compliance minimization problem for two different settings of boundary conditions**

## 3. Methodology

As briefly mentioned in the introduction, in the proposed method, the evaluation of the objective and/or the constrain in each TO iteration is conducted using MapNet instead of the numerical simulation used in the conventional TO methods. The main process of the evaluation can be separated into 5 steps. First, the fine-scale structure is scaled down to its coarse-scale structure (coarsening) and the FEM calculation is performed at the coarse scale to obtain the field needed for the evaluation of the objective and/or constraints. Next, the entire domain/field is decomposed into a set of small fragments (fragmentation).

A MapNet is then used to map the coarse-scale field of each fragment to the corresponding fine-scale field. At last, the fragments of the fine-scale field are combined to form the fine-scale field of the original domain (defragmentation). The process is illustrated in the flowchart shown in Fig. 3 with comparison to the conventional FEM-based TO process and the detailed description of each part is presented in the following paragraphs.

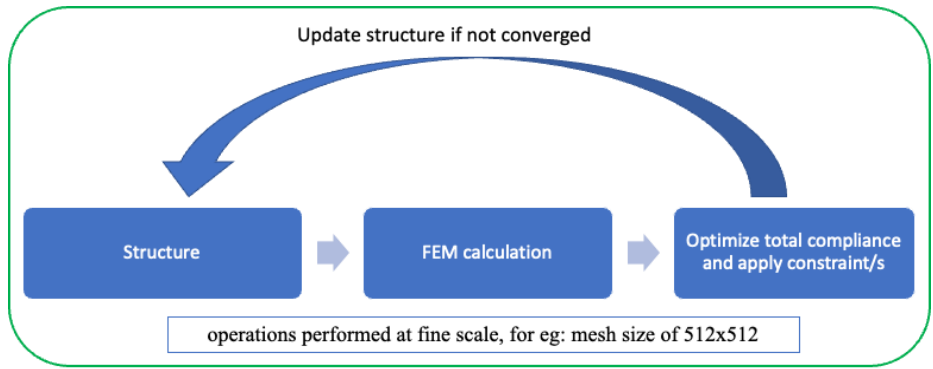

(a) Conventional FEM-based TO process

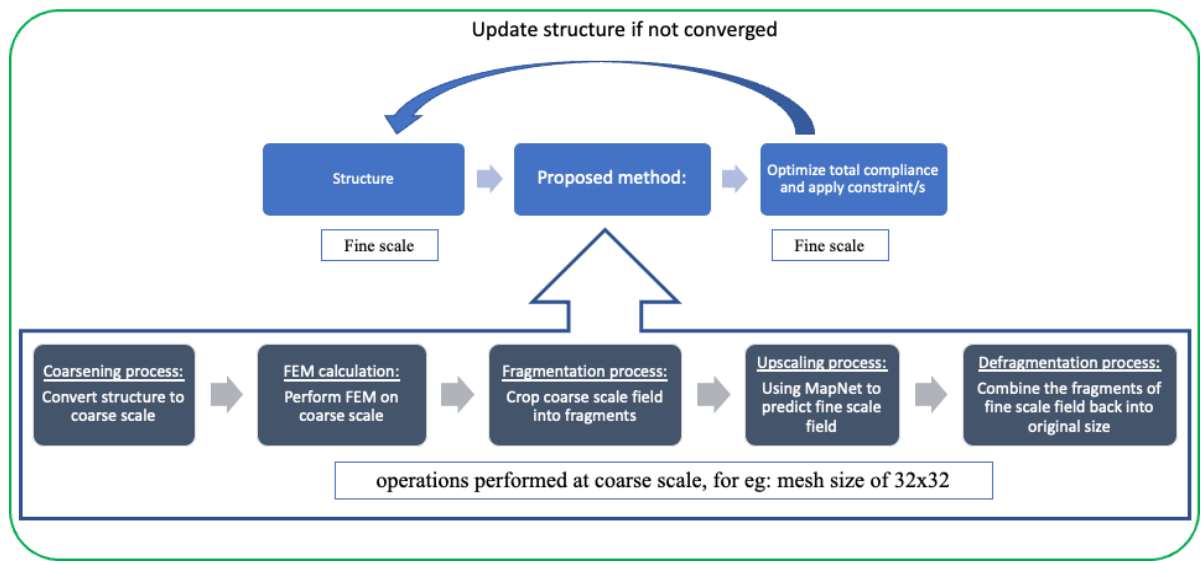

(b) MapNet-TO process

**Fig 3 Illustration of the conventional FEM-based TO (a) and the proposed framework based on MapNet (b)**

Coarsening is performed by scaling down the fine-scale density field of the structure with $N_F$ number of elements to its coarse-scale density field with $N_C$ number of elements, where $N_c$ should be much smaller than $N_F$ to achieve high efficiency. The density value of a coarse-scale element is obtained by averaging the density values of $N_F/N_c$ fine-scale elements. Fig 4 shows one sample of the conversion from a fine-scale structure to a coarse-scale structure. In the second step, FEM calculation is performed on the coarse-scale structure to obtain its physical field such as stress/strain field. Since the scale difference is large between the coarse and fine scale, the reduction in simulation time can be significant.

Next, fragmentation is performed on the entire field as illustrated in Fig 5. Using one sample of the density field obtained from the cantilever design problem as an example, the fine-scale structure is cropped into 64 small non-overlapping fragments. Moreover, instead of the non-overlapping cropping, we can go a step further by introducing overlapping cropping as shown in Fig 6. In fact, over-lapping cropping is more advantageous because when the fragments are combined to form the entire field, there is less likely for discontinuities to occur at the edge of each fragment, thus making the overall field smoother. More discussions will be presented in Section 4.1.3.

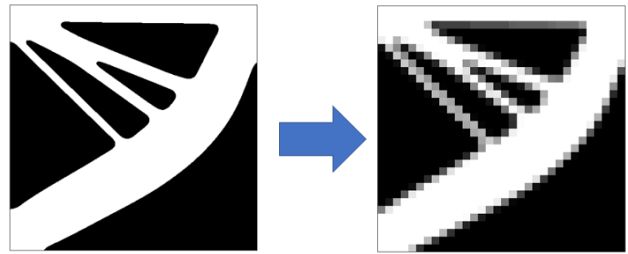

**Fig 4  Coarsening process to convert fine-scale structure to coarse-scale structure**

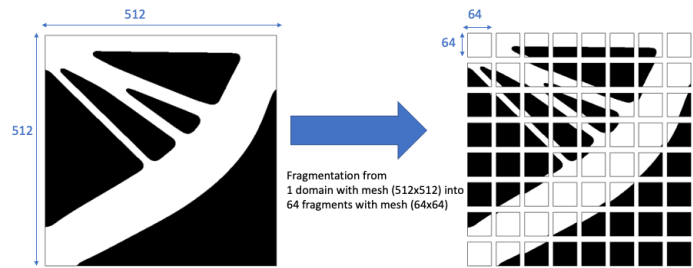

**Fig 5  Fragmentation process cropping the domain into small subdomains (fragments)**

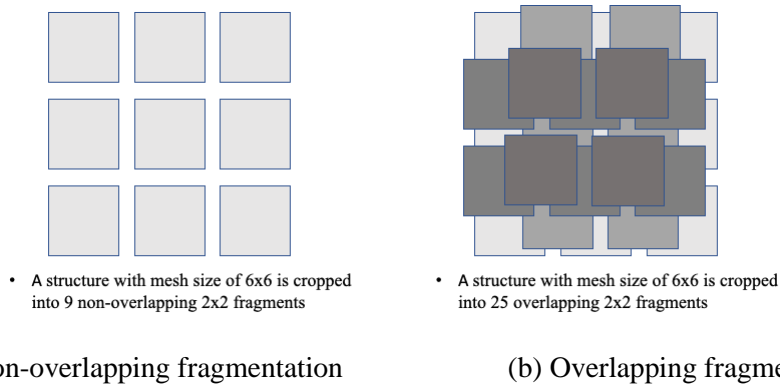

(a) Non-overlapping fragmentation        (b) Overlapping fragmentation

**Fig 6 Comparison between non-overlapping and overlapping fragmentation process**

As mentioned previously, using MapNet to map the field of a fragment instead of the whole domain helps improving the transferability. This is because fragments of different designs are more likely to resemble to each other even though the overall designs are entirely different. Considering the comparison between the cantilever beam and the L-shaped beam shown in Fig 7, it is not difficult to see that they are very different in view of the entire structure. However, as the domain is cropped into fragments, there are now more similar fragments as indicated by the connecting indicator in the figure. Therefore, if the MapNet is trained with the cropped cantilever beam data, it likely can provide accurate predictions for L-shaped beam without any retraining or retraining with very little data. The advantage is even more apparent in design cases with different domain sizes and shapes. Besides, the fragmentation process can also increase the number of training data for MapNet. For example, as demonstrated in Fig 5, 1 sample of structure has been cropped into 64 samples after the fragmentation, providing 64 times more samples that can be used to train MapNet. It should be pointed out that the fragmentation method is only possible with the special way MapNet is constructed. Since MapNet predicts fine-scale fields from coarse-scale fields, it does not need additional inputs about the external constraints on the problem, such as boundary conditions or loading conditions. However, if the ANN is built to map the density field of the structure to the corresponding mechanical field, fragmentation could not be performed because additional inputs on boundary/loading conditions are required.

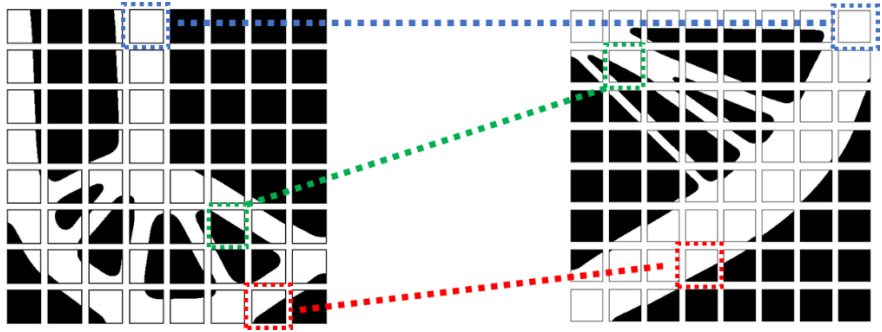

**Fig 7 Illustration showing similar fragments between the cropped L-shaped beam and cantilever beam structure**

After fragmentation, MapNet is used to map the coarse-scale field of each fragment to its fine-scale field. The architecture of the MapNet is illustrated in Fig 8, with each convolutional and deconvolutional layer having filter sizes of 3x3 and stride of 2x2 except for the last layer with 1x1 stride, and activation function of ReLU. The coarse-scale mesh size illustrated in the architecture of MapNet is chosen as 1/16$^{th}$ of the fine-scale mesh size in both width and height. In order to facilitate the training of MapNet, the data used as the input and output are all represented in 2D arrays, with each element in the array representing the field value at the corresponding location for the 2D design problem. However, due to the convolution-based architecture of MapNet, the dataset having N number of samples is reshaped into $(N \times W \times H \times 1)$, where W and H represents the width and the height of each array. We have found through experimentation that by applying the concept of U-Net into our MapNet, the performance can be improved. Another crucial feature that improves the prediction accuracy is by including the fine-scale structure at different deconvolutional layers.

To train the MapNet, we use the results obtained from early iterations of one FEM-based TO design. Specifically, we conduct a conventional FEM-based TO for the cantilever beam design problem described in Fig. 1 (a). The fine-scale strain energy field and density field generated during each iteration are extracted to form the dataset. Results from early iterations of TO are used as training data due to the reason that the topology or the density field of structure undergo major changes only during early iterations, while in late iterations only local fine tuning is involved [17]. The trained MapNet is then applied in other design problems. The fragments of fine-scale strain energy field predicted by MapNet are recombined back to the original scale through defragmentation process. For the non-overlapping defragmentation, the process is straightforward by just combining the prediction edge-to-edge. While for the overlapping case, each fragment is also combined but averaged values are taken at each overlapping area.

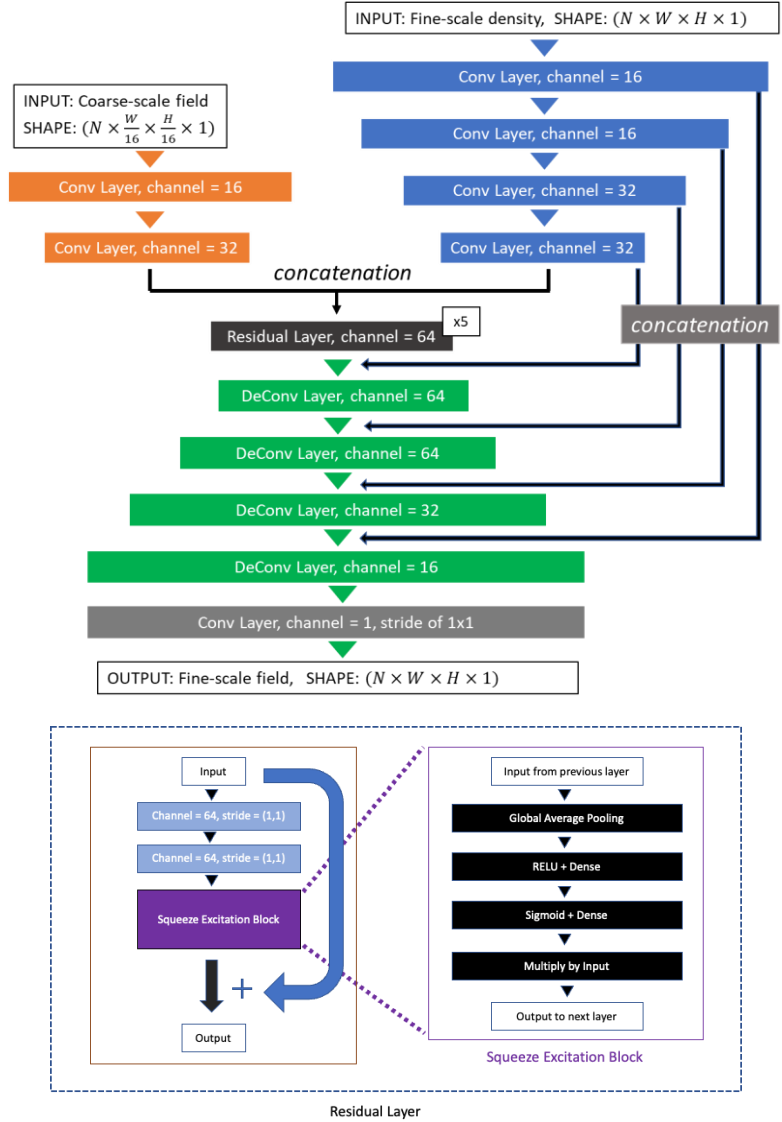

**Fig 8 The architecture of MapNet**

## 4. Results and discussions

In this section, the performance of the MapNet is firstly analysed. It is followed by the performance of the proposed method on several common 2D TO design problems.

### 4.1 Performance evaluation of MapNet

The cantilever beam design problem with a volume fraction constraint of 0.4 and a distributed load applied at the upper right corner as shown in Fig 1(a) is used to study the accuracy of the MapNet. A design resolution of $N_F = 512 \times 512$ is chosen as the fine-scale mesh, while the coarse-scale mesh is chosen to be $N_c = 32 \times 32$. The conventional FEM-based TO is run using BESO method to produce the design solution, which converges within 200 iterations. The data obtained from this TO process is used to train the MapNet following the procedure discussed in the previous section. First, the effect of the number of training data on the accuracy of the network is analysed. The performance of the MapNet is also compared to that without fragmentation and/or without embedding the fine-scale density in the network to demonstrate the effectiveness of the proposed method. Next, the fragmentation method is further analysed by investigating the effect of different fragment sizes on the performance. Lastly, the overlapping technique is discussed and used in the fragmentation process.

**4.1.1 Number of training data**

Three sets of fine-scale data with the total number of $N = 40, 60$ and $100$ are obtained by running the conventional TO process of the cantilever beam design for $N$ iterations. For example, to generate 40 fine-scale data, TO is only run up to 40 iterations. The coarse-scale strain energy field required for the training of the MapNet is obtained by performing FEA on the coarse-scale structure down scaled from the fine-scale structure following the method described in Section 3. The strain energy field is then cropped to small non-overlapping fragments using a cropping scale of 16, with each sample of the coarse-scale strain energy field is cropped from the original size of 32x32 to 256 samples of 2x2 non-overlapping fragments. The fine-scale density field and strain energy field are also cropped from their original size of 512x512 into 256 samples of 32x32 non-overlapping fragments. Hence the actual numbers of training data for the MapNet are $N_{train} = $ 256x40, 256x60 and 256x100 respectively in the three cases. The cropping process is shown in Fig 9. For better visualization, a cropping scale of 8 is used in the figure. Using these cropped data, the MapNet is trained using ADAM optimizer and a learning rate of 1e-4. Due to the limitation of computational power, the number of iterations is set to be 1000 steps. In order to ease the training process for neural network, the values of the input and output are normalized to around the range of 0 to 1. The normalization factors for the coarse and fine scale strain energy field are selected as 1e-4 and 1e-6 respectively, which are obtained from their distributions shown in Fig 10.

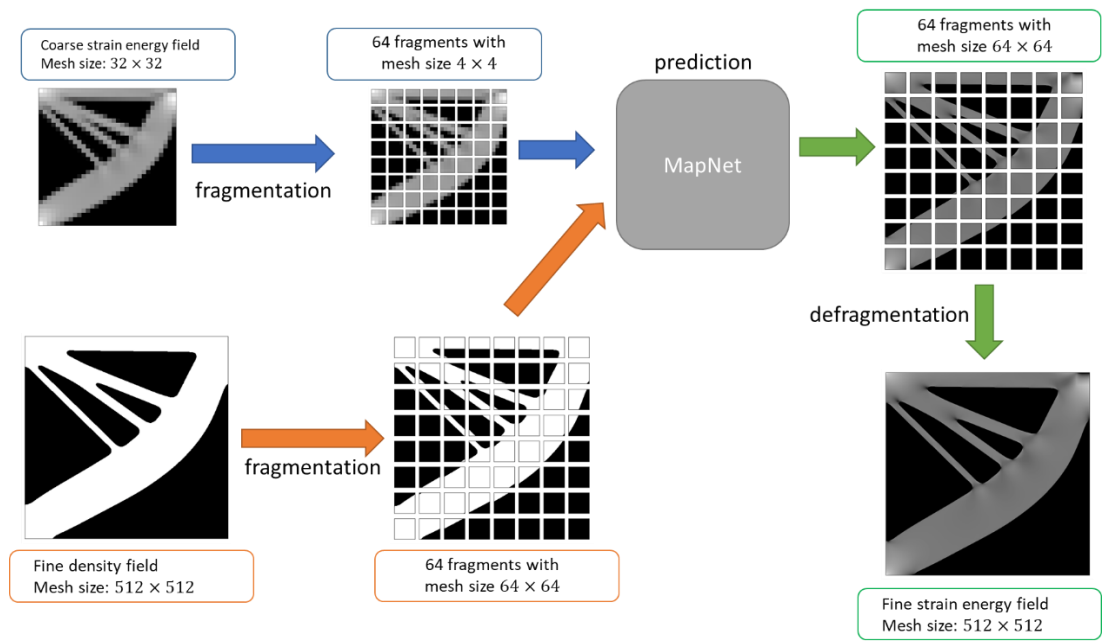

**Fig 9 Illustration showing the fragmentation, MapNet prediction and defragmentation process, starting from the coarse-scale strain energy field and fine-scale density field, finally to fine-scale strain energy field**

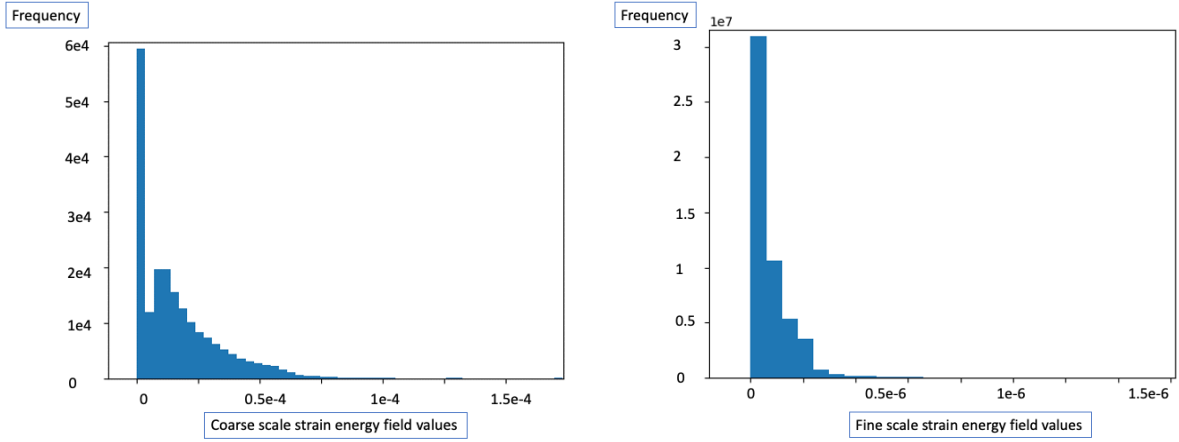

**Fig 10 Samples of the distributions of the coarse-scale and fine-scale strain energy fields obtained from the fragments of cantilever beam design problem**

After the MapNet is trained, 100 fine-scale data not including in the training dataset is used to evaluate the accuracy of the MapNet. In this case, the data from the 100th to the 200th iteration of the conventional FEM-based TO process are selected to be the testing data. Each data is cropped into 256 fragmented data. The fine-scale strain energy field predicted by the MapNet on each fragment is compared to that obtained through FEM (ground truth) directly obtained from the FEM-based TO and shown in Fig 11. For clear visualization, the strain energy field is plotted in the logarithm scale with an offset of 1e-8, that is, $U_{plot} = \log(U_{original} + 1e^{-8})$. To provide a quantitative error measure, the mean squared error (MSE) is calculated and listed in Table 1. The MSE is defined as $\frac{1}{N}\sqrt{\sum_{i=1}^{N}(U_i^{NN} - U_i^{FEM})^2}$, where $N$ is the total number of testing fragments, $U_i^{NN}$ and $U_i^{FEM}$ refer to the strain energy of the $i$th element predicted by the MapNet and the FEM respectively. From the figure and the table, it can be observed that as the number of training data increases, the MSE decreases and therefore the accuracy of the MapNet improves as expected. Based on the consideration of both accuracy and efficiency, the MapNet trained with data obtained from 60 TO iterations is selected and used in all the analyses presented in the rest of the paper. To demonstrate the effect of fragmentation and the inclusion of fine-scale density field in the neural network, the predicted fine-scale strain energy fields of fragments are also compared to those obtained from the MapNet trained without fragmentation, that is, the MapNet maps the coarse-scale field of 32x32 directly to the fine-scale field of 512x512, and to those obtained from the MapNet constructed without the inclusion of fine-scale density field. Results are presented in Fig 12. From the comparison, it is obvious that fragmentation and the inclusion of the fine-scale density greatly improve the prediction accuracy. In particular, adding the fine-scale density field produces results with much sharper and clearer edges. This is due to the reason that the fine-scale density field contains the information on the general shape of the structure and can provide a filtering effect on the predictions.

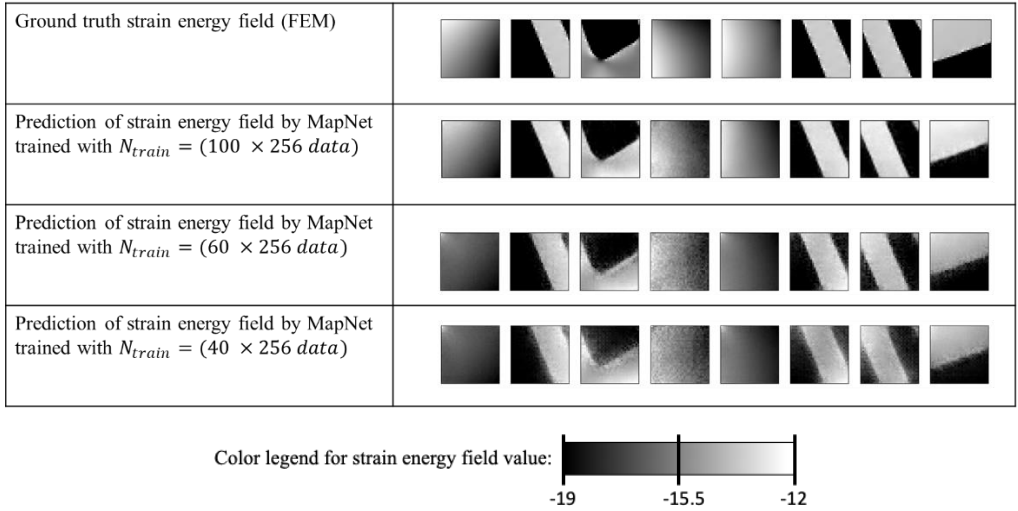

**Fig 11 Prediction of strain energy field (fragments) by the MapNet trained with the different number of training data samples (cantilever beam problem)**

**Table 1. Comparison of MSE between MapNet trained with different number of training data.**

| Number of training data | 100 | 60 | 40 |
|---|---|---|---|
| MSE | 0.00029 | 0.00058 | 0.00072 |

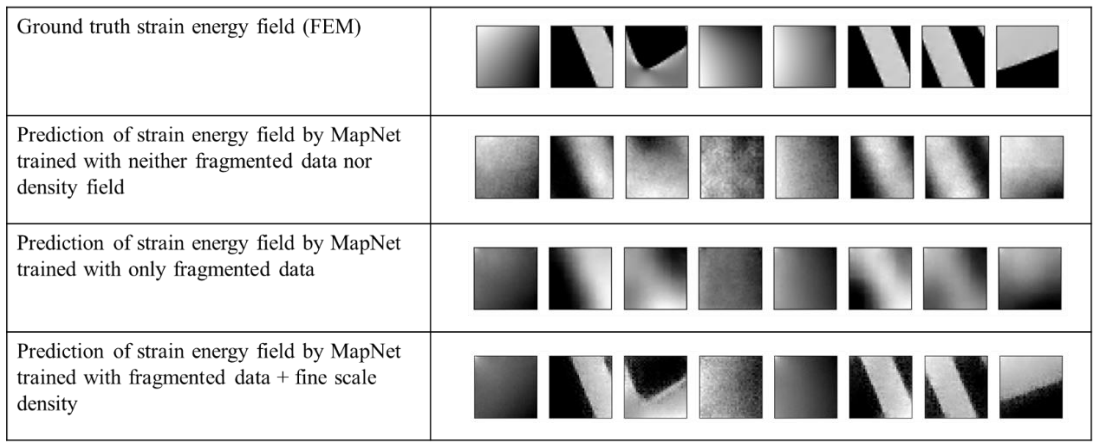

**Fig 12 Comparison of strain energy field of fragments predicted by MapNet trained with neither fragmented data nor density field, only fragmented data, and both fragmented data and density field to the ground truth. The colour legend is the same as that in Fig 11**

To study the performance of the MapNet, it is also important to examine the prediction accuracy of the entire field, that is, the field with its original size of 512x512 obtained after combining all fragmented fields with size of 32x32. This process, which is called defragmentation, is illustrated in Fig 9. The defragged strain energy field predicted by the MapNet trained with fragmented data and fine scale density is shown in the second row of Fig 13. Aside from visual comparison, a quantitative comparison is also made by calculating the mean squared error of the prediction, which is tabulated in Table 2. The predictions by the MapNet without fragmentation and the density field is also shown in the same figure and table for comparison. It can be observed that the predicted fine-scale strain energy field with our proposed method is also much better in the defragged form.

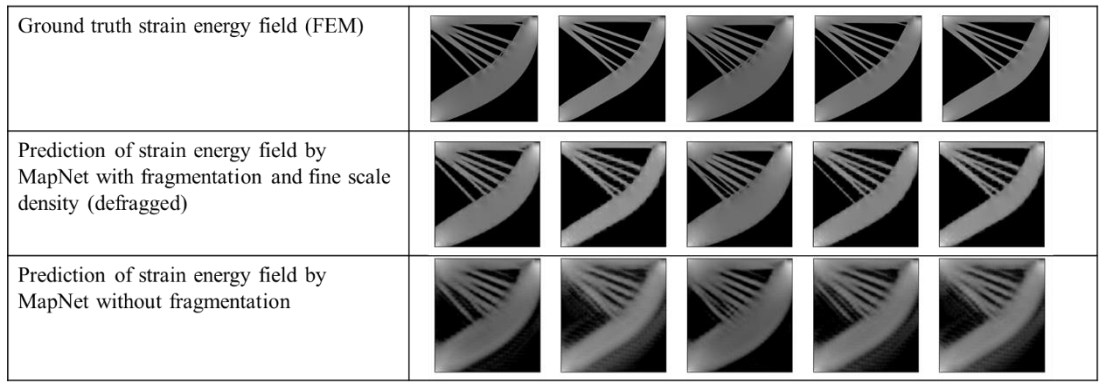

**Fig 13 Comparison between predicted strain energy field by MapNet trained with neither fragmentation nor fine-scale density, and that trained with both**

**Table 2. Comparison between MSE of the network trained with neither fragmentation nor fine-scale density, and that trained with both.**

| **MapNet** | Trained without fragmentation and fine-scale density | Trained with fragmentation and fine scale density |
|---|---|---|
| **MSE** | 0.0017 | 0.0006 |

### 4.1.2 Fragmentation – cropping scale

Results shown in the previous section indicate that fragmentation improves the prediction accuracy of the neural network. In these results, a cropping scale of 16 is used for the fragmentation, which has increased the training data by 256 times. If a larger cropping scale is used, for example with a scale of 32, the number of training data can be further increased, and consequently, the accuracy of the MapNet would be further improved. However, our study indicates that this is not necessarily the case. The investigation is performed by training the MapNet with three different cropping scales of 8, 16 and 32 respectively. Results in the defragged form are compared in Fig 14 along with the prediction error listed in Table 3 for all the cases.

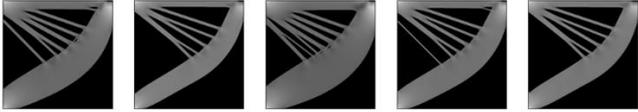

(a)

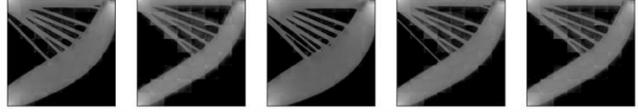

(b)

**Fig 14 (a) Comparison of strain energy field predicted by MapNet trained with fine-scale density and different cropping scales of fragments. (b) Zoomed in comparison of the predicted strain energy field from results in the first two columns in figure (a)**

**Table 3. Comparison of MSE from the MapNet trained with different cropping scale of fragments**

| Cropping scale | 8 | 16 | 32 |
|---|---|---|---|
| MSE | 0.00088 | 0.00058 | 0.00152 |

By comparing the results from the cropping scale of 8 with that of 16, it can be observed that the MSE of the MapNet decreases with the cropping scale. This is due to the previous explanation that as the cropping scale increases, more fragments are produced, thus increasing the number of training data for the MapNet. However, as the scale increases to 32, which corresponds to a fragment size of 1x1 for the coarse-scale data, and 16x16 for fine-scale data, the performance dropped with a higher error. One reason is due to the non-smooth boundaries between fragments as can be observed in Fig 14(b). Since the fine-scale strain energy of the entire field is obtained by simple recombination of all non-overlapping fragments, it is to be expected that the value at the edges of fragments might not be continuous with the adjacent fragments. Another reason might be due to that as the samples get cropped into smaller fragments, the possibility for non-uniqueness to occur increases. Non-uniqueness refers to the case where the same input for the network corresponds to different output values, essentially having a one-to-many mapping, which makes the training of the network more difficult. For example, as shown in Fig 10, the coarse-scale strain energy ranges from 0 to 1.5e-4. Therefore, any two values having difference smaller than 1e-12 are regarded as being identical. By using this criterion and considering the fragmented data for the cropping scale of 32 with a fragment size of 1x1, it has been found that two fragments with the same coarse-scale strain energy value of 2.57e-5 and the exactly identical density field, have two different fine-scale strain energy fields as illustrated in Fig 15. Therefore, considering the trade-off, the cropping scale should be too large so that the non-uniqueness issue could be avoided during the training of the MapNet. In this case, the best cropping scale is 16.

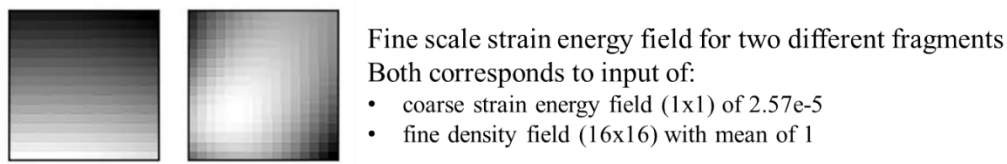

Fig 15 Illustration showing examples of non-uniqueness issue

### 4.1.3 Fragmentation – overlapping fragmentation

Although the results obtained by fragmentation are already much better than those without fragmentation, by careful observation it can be found that the predicted fine-scale strain energy field in the defragged form is not very smooth at the boundaries of two fragments due to the reason explained in the previous section. In order to overcome this issue, the overlapping method described in the methodology section (Fig 6) is utilized. Referring back to the figure, the overlapping is performed during the fragmentation by cropping the domain at a smaller interval. Considering the original domain of 32x32, with a cropping scale of 16, the domain is cropped at each interval of 2 elements. With overlapping, the domain is instead cropped at each interval of 1 element. Therefore, a side length of 32 elements will be cropped into 31 fragments instead of the previous 16. Similarly, after the fragments of the fine-scale strain energy are predicted by MapNet, the fragments are also combined at the interval of 16 elements, with average values taken at any overlapping parts. The results obtained with overlapping fragments are shown in Fig 16 with the MSE tabulated in Table 4. By comparing to the previous result with non-overlapping fragmentation in the same figure and table, the predicted fine-scale strain energy field is observed to be smoother and the error of the MapNet has also decreased further with the utilization of overlapping fragments.

| Ground truth strain energy field (FEM) | |
|---|---|
| Strain energy field predicted by MapNet trained with overlapping fragmentation | |
| Strain energy field predicted by MapNet trained with non-overlapping fragmentation | |

(a) Strain energy field predicted by MapNet with and without using overlapping fragmentation.

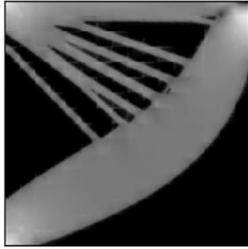

(b) Zoomed in view of predicted strain energy field for the first column of result in figure (a)

**Fig 16 Comparison between strain energy field predicted by MapNet trained with non-overlapping and overlapping fragmentation**

**Table 4. Comparison of the MSE between MapNet trained with overlapping fragmentation and non-overlapping fragmentation.**

| Number of Training Data ($N_{train}$) | Overlapping fragmentation | Non-overlapping fragmentation |
|---|---|---|
| **MSE** | 0.0004 | 0.0006 |

### 4.2 Application of the MapNet to TO design

With the MapNet developed and trained using the method discussed in the previous section, it is then implemented into the TO process and the TO process for the cantilever design with a single load illustrated in Fig 1 (a) is carried out following the process introduced in the methodology section. In this case, BESO algorithm is used. For clarity, it should be pointed out that the MapNet used in all structure designs presented in the rest of the paper is trained with only 60 TO data obtained from the first 60 TO iterations of the cantilever beam design shown Fig 1 (a) , and with a cropping scale of 16 and the over-lapping fragmentation. The optimized structure obtained from this modified TO process, that is, the MapNet-based TO process, is compared to that obtained from the conventional FEM-based TO process in Fig 17. The objective function, that is, the compliance associated with both structures is also provided in the figure. In addition, the optimized structure obtained from the MapNet constructed without fragmentation and the inclusion of the fine-scale density field is also shown to demonstrate the superiority of the current MapNet architecture. From the comparison, the first thing to note is that the result from MapNet implementing fragmentation and with the fine-scale density is obviously much better than that without. At the same time, the optimized structure obtained with the MapNet-TO is very similar to that obtained using the conventional method.

Table 5 shows the detailed breakdown of the time for the relevant phases in each case and the total time required for each iteration in the TO process. The computational time shown is based on the calculation performed on Intel(R) Xeon(R) CPU E5-2687Wv2 (3.40GHz) containing 16 cores. It should also be mentioned that the FEM calculations in this work are performed using the sparse linear system solver in Python's SciPy library due to the performance and memory efficiency provided by the library. From the table, it can be observed that a 300-times reduction in the computational time is achieved with the MapNet-based TO in one single TO iteration. Considering the fact that the TO design in this cantilever beam case requires 200 iterations to converge and only the first 60 iterations are used to generate the training data, the total time saving for completing even one TO process is significant.

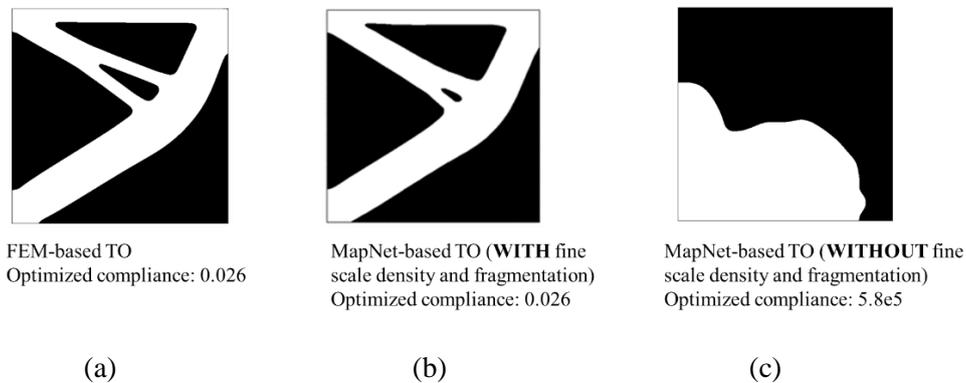

(a) FEM-based TO
Optimized compliance: 0.026

(b) MapNet-based TO (**WITH** fine scale density and fragmentation)
Optimized compliance: 0.026

(c) MapNet-based TO (**WITHOUT** fine scale density and fragmentation)
Optimized compliance: 5.8e5

**Fig 17 TO results of the cantilever beam problem with the load applied at the top right boundary obtained from (a) FEM-based TO method; (b) MapNet-based TO method; (c) MapNet-based TO method fragmentation and the inclusion of fine-scale density**

**Table 5. Time comparison for completion of a single TO iteration by pure FEM-based TO method and the proposed MapNet-based TO method.**

| Time category | FEM-TO process | MapNet-TO process |
|---|---|---|
| FEM calculation time (s) | ~ 600 (Mesh 512x512) | ~ 0.13 (Mesh 32x32) |
| MapNet calculation time + fragmentation and defragmentation time (s) | 0 | 0.92 |
| The time for the rest of optimization process (s) | ~1 | ~1 |
| Total time for each iteration (s) | ~ 600 | ~ 2 |

## 4.3 Transferability of the MapNet

In this section, the transferability of the MapNet is demonstrated on several benchmark design problems that are different from the cantilever design case used to train the MapNet. They are the cantilever beam design with multiple applied forces, the L-shaped beam design as well as the bridge design problem illustrated in Fig 1. Specifically, the MapNet trained using the cantilever design with a single load is directly implemented into the TO process to solve the three new design problems without any re-training.

The first two design problems shown in Fig 1(b) and Fig 1(c) have the same square design domains with the fine-scale mesh of 512x512 and the coarse-scale mesh of 32x32. BESO algorithm is used as the TO method for all three design cases. The optimized structures and their final compliances obtained through the conventional FEM-based TO and the MapNet-based TO are shown in Fig 18 and Fig 19. As indicated by the results, the optimized structures obtained using the MapNet-based TO method resemble those from the FEM-based TO method with fewer branches. The final compliances obtained are also similar to those from FEM-based TO.

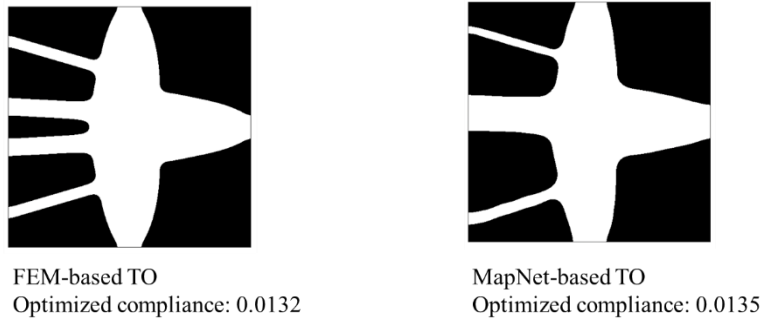

FEM-based TO
Optimized compliance: 0.0132

MapNet-based TO
Optimized compliance: 0.0135

**Fig 18 Comparison of TO results between FEM-based method and the proposed MapNet-based method for cantilever beam with multiple loads applied**

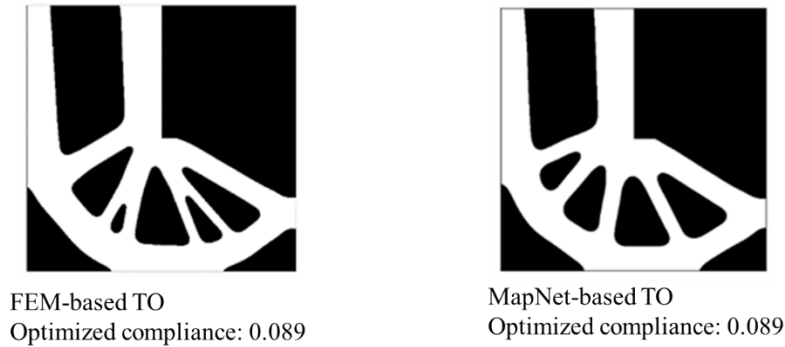

FEM-based TO
Optimized compliance: 0.089

MapNet-based TO
Optimized compliance: 0.089

**Fig 19 Comparison of TO results between FEM-based method and the proposed MapNet method for L-shaped beam**

The bridge design problem illustrated in Fig 1(d)) has a very different boundary condition and domain shape and size from all other design cases considered. The design domain is discretized into 768(width) x 384(height) elements. In this case, the coarse-scale mesh is chosen to be $N_c = 48 \times 24$. During the fragmentation process, each sample of the coarse-scale strain energy field is cropped into $47 \times 23$ fragments of $2 \times 2$ with overlapping cropping of 1 element. The fine-scale density field is cropped into $47 \times 23$ fragments of size $32 \times 32$. These inputs are then fed to the MapNet to predict the fine-scale strain energy fields of all $47 \times 23$ fragments with size of $32 \times 32$. These fine-scale strain energy fields are then combined to form the entire field of the original domain, that is, with the size of $768 \times 384$, and the TO process is proceeded as previously discussed. Although the design domain of the bridge is entirely different from that of the cantilever, by using the fragmentation process, the previously trained MapNet can still be directly applied to this problem because it provides the prediction on the fragments/building blocks of the structure instead of the whole structure. Therefore, its generalization capability is much increased. The optimized results obtained from the MapNet-based TO process and FEM-based TO method are shown in Fig 20. From the results, the MapNet trained with the cantilever beam data again shows excellent performance with the optimized structure and its compliance being very close to that of the ground-truth results.

It should be pointed out that the strain energy field needs to be properly normalized before feeding it into the MapNet, otherwise the predicted fine-scale strain energy field would have a large error. Since the ranges of the strain energy of different problems can be different, different normalization factors should be determined and used for different problems. This can be done by observing the coarse and fine-scale strain energy fields from the first several iterations, for example, 5 iterations of the FEM-based TO process. In the first two design cases, the normalization factor is the same as the simple cantilever design with a single load. In the bridge case, it is found that the normalization factors should be set as 1e-7 and 1e-9 for the coarse and fine-scale data respectively.

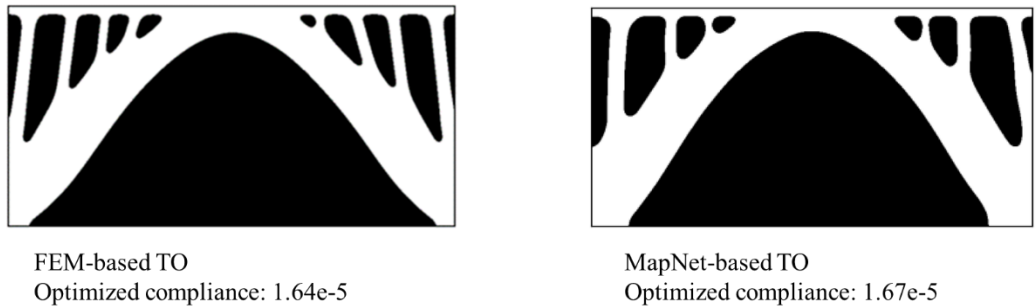

FEM-based TO
Optimized compliance: 1.64e-5

MapNet-based TO
Optimized compliance: 1.67e-5

**Fig 20 Comparison of TO results between FEM-based method and the proposed MapNet method for the bridge design problem**

To examine the efficiency, the computational times required to perform the MapNet-based TO in the three new design problems are calculated. For the cantilever beam with multiple applied loads and the L-shaped beam, the time saving per design iteration is similar to that tabulated in

Table 5 because they have the same mesh size as the previous cantilever beam problem (512x512). The time saving per iteration for the bridge design problem is slightly different, which is around 250 times as shown in Table 6.

**Table 6. Comparison between computational time for one single TO iteration for FEM-based method and the proposed MapNet method for the bridge design problem.**

| Time category | TO process with pure FEM calculations on fine scale | TO process with MapNet implementation |
|---|---|---|
| FEM calculation time (s) | ~ 750 (Mesh 768x384) | ~ 0.17 (Mesh 48x24) |
| MapNet calculation time + fragmentation and defragmentation time (s) | 0 | ~2 |
| The time for the rest of optimization process (s) | ~1 | ~1 |
| Total time for each iteration (s) | ~ 750 | ~ 3 |
| Total time taken for whole TO process (assuming 200 iterations for convergence) | $1.5 \times 10^5$ | $6 \times 10^2$ |

### 4.4 Applications of the MapNet with SIMP

In this section, the proposed method is implemented in the SIMP-based TO process to show that it is not restricted by the type of TO methods. In the SIMP approach, the density field contains intermediate values, and thus a new MapNet is constructed. Following the procedure described in previous sections and again using the cantilever beam design with a single load to generate training data, the MapNet is found to be requiring a minimum of 60 training data from the TO process in order to achieve satisfactory prediction accuracy. The performance of the MapNet is then evaluated by directly implementing it into the SIMP process and comparing the design solution with that obtained from FEM-based SIMP on this design case. Satisfactory results are obtained as shown in Fig 21.

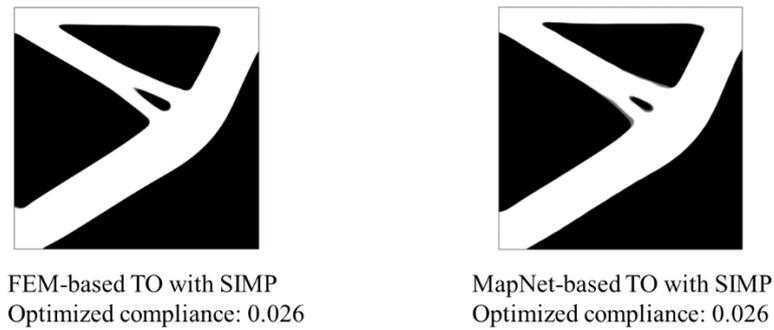

FEM-based TO with SIMP
Optimized compliance: 0.026

MapNet-based TO with SIMP
Optimized compliance: 0.026

**Fig 21 Comparison between TO results of FEM-based method and the proposed MapNet method (SIMP) for cantilever beam with a single applied load at the top of the right boundary**

Next for the demonstration of transferability, the MapNet-based SIMP is used directly to design the L-shaped beam and the bridge. The optimized results obtained for each corresponding design problem are compared to the design solutions obtained from the FEM-based SIMP in Fig 22. These results again illustrate the good transferability of the proposed method.

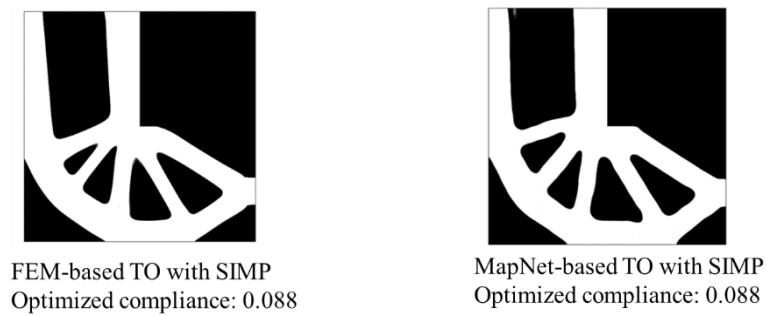

FEM-based TO with SIMP
Optimized compliance: 0.088

MapNet-based TO with SIMP
Optimized compliance: 0.088

(a) L-shaped beam design problem

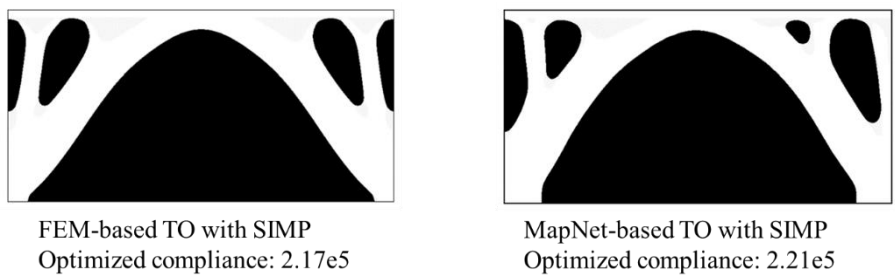

FEM-based TO with SIMP
Optimized compliance: 2.17e5

MapNet-based TO with SIMP
Optimized compliance: 2.21e5

(b) Bridge design problem

**Fig 22 Comparison between TO results of FEM-based method and the proposed MapNet method (SIMP) for (a) L-shaped beam and (b) bridge design problem**

### 4.5 Application to the thermal problem

Similar to the FEM-based TO method, the proposed MapNet-based TO method is applicable to a wide range of design problems. In this section, results of the structure design with minimum thermal compliance are shown to further demonstrate the performance of the MapNet-based TO method. The two thermal design problems are described in Section 2. The first design problem being considered has a small size of heat sink with a constant temperature of zero degree located at the centre of the top boundary (Fig 2(a)). The domain size of the design problem is again selected to be 512x512, with the volume fraction constraint chosen to be 0.4. This problem is firstly solved using the FEM-based SIMP, the filter radius is selected to be 16. The process converges around 100 iterations and the optimized results are shown in the left figure of Fig 23. A new MapNet with the same architecture as the previous one is constructed and trained using the data obtained from the first few iterations of the FEM-based TO. Specifically, it is found that 40 TO data is sufficient to train the MapNet to satisfactory performance. The coarse-scale mesh size for this problem is again 32x32. By observing the distribution and range of thermal compliances from the first 5 iterations of FEM-based TO, the normalization factors for this thermal problem are selected to be 200 and 5 for the coarse and fine-scale thermal compliance respectively. Following the same approach shown in Fig 3, the trained MapNet is implemented into the TO process and the thermal compliance is minimized using the MapNet-based TO. The optimized results are shown in Fig 23 and it is observed that the results are close to those from the FEM-based TO.

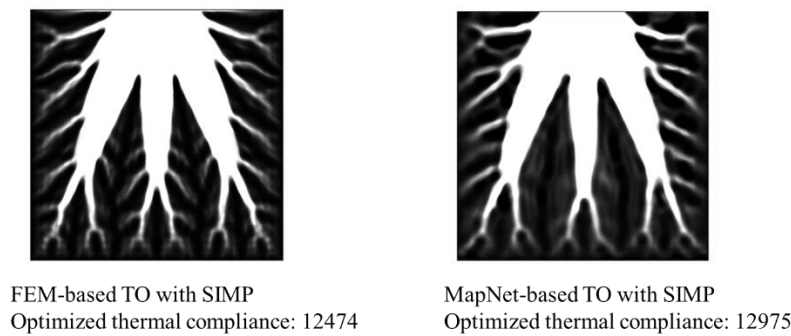

FEM-based TO with SIMP  
Optimized thermal compliance: 12474

MapNet-based TO with SIMP  
Optimized thermal compliance: 12975

**Fig 23 Comparison between TO results of FEM-based method and the proposed MapNet method for thermal problem with a small heat sink**

To demonstrate the transferability of the trained MapNet, the thermal design problem with a large heat sink located on the top of the boundary (Fig 2(b)) is solved using the previously trained MapNet. Instead of imposing the same volume fraction of 0.4 as the previous problem, we go a step further by setting a different volume fraction constraint for this problem, which is 0.6. The design result using MapNet is presented in Fig 24 together with that obtained from the FEM-based TO. Although both the volume fraction constraint and boundary condition are different from the first problem, thus ending up with an optimized structure which looks very different, the MapNet-based method can still provide a result closely resembling that by FEM. This result again demonstrates the good transferability of the MapNet.

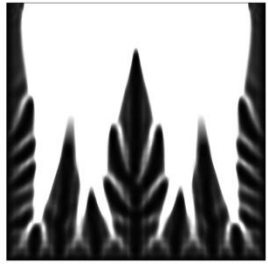 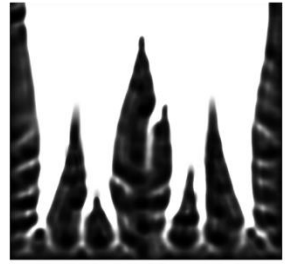

FEM-based TO with SIMP  
Optimized thermal compliance: 4943

MapNet-based TO with SIMP  
Optimized thermal compliance: 5032

**Fig 24 Comparison between TO results of FEM-based method and the proposed MapNet method for thermal problem with a large heat sink and a volume fraction of 0.6**

## 5. Conclusion

In this work, an adaptive and scalable deep learning-based method is proposed to speed up the iterative TO design process with large design domain. The time-consuming calculation of the field of interest is replaced with that performed at a much coarse mesh and a deep learning model known as MapNet is developed to map the coarse field back to the fine field. A unique feature of the MapNet is that it is constructed for the building blocks of the domain instead of the entire domain. As such, the MapNet can be applied to problems with different domain sizes and shapes without retraining. The performance of the proposed method is demonstrated across different design problems frequently used as benchmark in structure design. The MapNet is shown to be able to provide predictions on the fine-scale strain energy field with only a small amount of training data obtained from the first few iterations of TO process of one cantilever beam design problem. By implementing the trained MapNet into the TO process, the optimized results of all three different testing design problems are shown to be closely similar to those obtained from conventional FEM-based TO. In the meantime, the total computational time required for each iteration has been greatly reduced.

We have shown that the proposed MapNet can be implemented into both BESO and SIMP. We have also shown that the method can be used for thermal design problems. In fact, it could be implemented into any density-based TO methods and applied to a wide range of design problems including nonlinear problems and stress-constrained designs similar to the conventional FEM-based TO method. To extend the proposed method to these problems will be our near-future work.

Although only 2D design problems are demonstrated in the current work, the method could be developed for 3D designs and it will be carried out in our future work. As the computational time involved in 3D simulations are much longer than that in 2D, the potential time saving would be massive if a large-scale difference is chosen between the coarse and fine scale. Another investigation that could also be studied in future work is the normalization factor for the inputs and outputs of the network. In the current work the normalization factors are chosen so that the distribution of the strain energy field is scaled to match that of the training data. This requires some prior knowledge or observations to be made on some of the data from the design problem that are being considered. Better normalization methods could be developed to minimize the requirement of prior knowledge on the design problems.

**Data Availability**

The codes used for the implementation of the proposed method including architecture of MapNet and topology optimization algorithms are available upon reasonable request.

**References**

[1] N. Aage, E. Andreassen, B. S. Lazarov and O. Sigmund, (2017) Giga-voxel computational morphogenesis for structural design. *Nature.* vol. 550, *(7674),* pp. 84-86.

[2] J. Yvonnet and Q. He, (2007) The reduced model multiscale method (R3M) for the non-linear homogenization of hyperelastic media at finite strains. *Journal of Computational Physics.* vol. 223, *(1),* pp. 341-368.

[3] N. C. Nguyen, (2008) A multiscale reduced-basis method for parametrized elliptic partial differential equations with multiple scales. *Journal of Computational Physics.* vol. 227, *(23),* pp. 9807-9822.

[4] S. Boyaval, (2008) Reduced-basis approach for homogenization beyond the periodic setting. *Multiscale Modeling & Simulation.* vol. 7, *(1),* pp. 466-494.

[5] E. Monteiro, J. Yvonnet and Q. He, (2008) Computational homogenization for nonlinear conduction in heterogeneous materials using model reduction. *Computational Materials Science.* vol. 42, *(4),* pp. 704-712.

[6] M. Cremonesi, D. Néron, P. Guidault and P. Ladevèze, (2013) A PGD-based homogenization technique for the resolution of nonlinear multiscale problems. *Comput. Methods Appl. Mech. Eng.* vol. 267, pp. 275-292.

[7] J. A. Hernández, J. Oliver, A. E. Huespe, M. A. Caicedo and J. Cante, (2014) High-performance model reduction techniques in computational multiscale homogenization. *Comput. Methods Appl. Mech. Eng.* vol. 276, pp. 149-189.

[8] P. Benner, S. Gugercin and K. Willcox, (2015) A survey of projection-based model reduction methods for parametric dynamical systems. *SIAM Rev.* vol. 57, *(4),* pp. 483-531.

[9] D. A. White, W. J. Arrighi, J. Kudo and S. E. Watts, (2019) Multiscale topology optimization using neural network surrogate models. *Comput. Methods Appl. Mech. Eng.* vol. 346, pp. 1118-1135.

[10] R. K. Tan, N. L. Zhang and W. Ye, (2020) A deep learning–based method for the design of microstructural materials. *Structural and Multidisciplinary Optimization.* vol. 61, *(4),* pp. 1417-1438.

[11] Z. Nie, H. Jiang and L. B. Kara, (2020) Stress field prediction in cantilevered structures using convolutional neural networks. *Journal of Computing and Information Science in Engineering.* vol. 20, *(1),* pp. 011002.

[12] S. Lee, H. Kim, Q. X. Lieu and J. Lee, (2020) CNN-based image recognition for topology optimization. *Knowledge-Based Syst.* vol. 198, pp. 105887.

[13] K. A. Kalina, L. Linden, J. Brummund, P. Metsch and M. Kästner, (2022) Automated constitutive modeling of isotropic hyperelasticity based on artificial neural networks. *Comput. Mech. vol 69,* pp. 213-232.

[14] I. Sosnovik and I. Oseledets, (2019) Neural networks for topology optimization. *Russian Journal of Numerical Analysis and Mathematical Modelling.* vol. 34, *(4),* pp. 215-223.

[15] Y. Yu, T. Hur, J. Jung and I. G. Jang, (2019) Deep learning for determining a near-optimal topological design without any iteration. *Structural and Multidisciplinary Optimization.* vol. 59, *(3),* pp. 787-799.

[16] H. T. Kollmann, D. W. Abueidda, S. Koric, E. Guleryuz and N. A. Sobh, (2020) Deep learning for topology optimization of 2D metamaterials. *Mater Des.* vol. 196, pp. 109098.

[17] G. C. Ates and R. M. Gorguluarslan, (2021) Two-stage convolutional encoder-decoder network to improve the performance and reliability of deep learning models for topology optimization. *Structural and Multidisciplinary Optimization.* vol. 63, *(4),* pp. 1927-1950.

[18] C. Qian and W. Ye, (2021) Accelerating gradient-based topology optimization design with dual-model artificial neural networks. *Structural and Multidisciplinary Optimization.* vol. 63, *(4),* pp. 1687-1707.

[19] M. Baandrup, O. Sigmund, H. Polk and N. Aage, (2020) Closing the gap towards super-long suspension bridges using computational morphogenesis. *Nature Communications.* vol. 11, *(1),* pp. 1-7.

[20] H. Chi, Y. Zhang, T. L. E. Tang, L. Mirabella, L. Dalloro, L. Song and G. H. Paulino, (2021) Universal machine learning for topology optimization. *Comput. Methods Appl. Mech. Eng.* vol. 375, pp. 112739.

[21] M. Raissi, P. Perdikaris and G. E. Karniadakis, (2019) Physics-informed neural networks: A deep learning framework for solving forward and inverse problems involving nonlinear partial differential equations. *Journal of Computational Physics.* vol. 378, pp. 686-707.

[22] L. Lu, X. Meng, Z. Mao, and E. George K. (2020) DeepXDE: A deep learning library for solving differential equations. *SIAM Review*, vol. 63, No. 1, pp. 208-228.

[23] M. Raissi and G. E. Karniadakis, (2018) Hidden physics models: Machine learning of nonlinear partial differential equations. *Journal of Computational Physics.* vol. 357, pp. 125-141.

[24] M. Raissi, A. Yazdani and G. E. Karniadakis, (2020) Hidden fluid mechanics: Learning velocity and pressure fields from flow visualizations. *Science.* vol. 367, *(6481),* pp. 1026-1030.

[25] D. Wang, C. Xiang, Y. Pan, A. Chen, X. Zhou and Y. Zhang, (2021) A deep convolutional neural network for topology optimization with perceptible generalization ability. *Engineering Optimization.* pp. 1-16.

[26] Ren Kai Tan, Q. Chao, Michael Wang and Wenjing Ye, (2022) An efficient data generation method for ANN-based surrogate models. *Structural and Multidisciplinary Optimization.* Vol. 65, No. 3, pp. 1-22.

[27] M. P. Bendsoe and O. Sigmund, (2003) *Topology Optimization: Theory, Methods, and Applications. Springer Science & Business Media.*